\pdfoutput=1
\documentclass[runningheads,orivec]{llncs}
\usepackage[T1]{fontenc}
\usepackage{lmodern}
\usepackage{graphicx,booktabs,array,multirow,amsmath,amssymb,adjustbox,placeins,float,xspace}
\usepackage{microtype}
\usepackage[hidelinks]{hyperref}
\newcommand{\pp}{\nobreak\hspace{0.16667em}\textnormal{pp}\xspace}

\newcommand{\tighttable}{\small\setlength{\tabcolsep}{4pt}\renewcommand{\arraystretch}{1.18}}
\newcommand{\tinytab}{\footnotesize\setlength{\tabcolsep}{3.5pt}\renewcommand{\arraystretch}{1.18}}
\newcommand{\panel}[1]{\par\vspace{6pt}\textbf{#1}\par\vspace{6pt}}
\newcommand{\ind}{\operatorname{I}}
\newcommand{\SupplementaryArchiveURL}{https://doi.org/10.5281/zenodo.23970406}

\hypersetup{pdftitle={When Audio and Video Mislead: Evaluating Multimodal AI for Conversational Emotion Recognition},pdfauthor={Youngseok Kim, Minjae Kim, Eunsoo Lee, Jihie Kim}}
\begin{document}
\raggedbottom
\clubpenalty=10000
\widowpenalty=10000
\brokenpenalty=10000
\title{When Audio and Video Mislead: Evaluating Multimodal AI for Conversational Emotion Recognition}
\titlerunning{When Audio and Video Mislead}
\author{Youngseok Kim \and Minjae Kim \and Eunsoo Lee \and Jihie Kim\thanks{Corresponding author.}}
\authorrunning{Y. Kim et al.}
\institute{Dongguk University, Seoul, Republic of Korea\\
\email{hhon0213@dgu.ac.kr, kim0729@dgu.ac.kr}\\
\email{dmstn7432@dgu.ac.kr, jihie.kim@dgu.edu}}
\maketitle
\begin{abstract}
Multimodal Emotion Recognition in Conversation (MERC) combines lexical, acoustic, visual and conversational cues, but accepting audio and video does not establish their effective or robust use. We audit off-the-shelf Omni models while holding the target transcript and prior textual context fixed and adding, removing or replacing target audio and video. In the primary evaluation of Qwen2.5-Omni-7B, Baichuan-Omni-1.5 and MiniCPM-o 2.6 on MELD and the full five-session six-class IEMOCAP set, T+A improves Accuracy in all six model--dataset pairs, with four gains significant after false discovery rate (FDR) correction. Across three source mappings, unrelated A/V (RandA/V) lowers Accuracy relative to Full in all 18 comparisons; all 18 remain significant under dialogue-cluster inference and BH-FDR correction, both on common-valid pairs and when invalid/OOM outcomes are treated as incorrect. Comparisons with modality-omission baselines further distinguish replacement penalties from mismatch-specific harm. A Qwen/MELD case study finds significantly greater gold-vs-best-candidate margin recovery in Late than Early or Middle under decision-query audio knockout, relative to an edge-count-matched Video-Sham control, using Holm-corrected tests. These results distinguish genuine-modality benefit from sensitivity and vulnerability to misleading input.
\end{abstract}

\section{Introduction}
Human emotion in conversation is conveyed through language, prosody, voice quality, facial expression and dialogue context. Multimodal Emotion Recognition in Conversation (MERC) combines these signals, commonly using MELD and IEMOCAP~\cite{busso2008iemocap,poria2019meld}. Recent systems model modality-specific evidence and conversational structure~\cite{li2023revisiting,seong2026emotion}, while Omni models can accept raw or near-raw text, audio and video within an integrated model~\cite{li2025baichuan,xu2025qwen}. Yet accepting a modality does not establish its effective use: a model may ignore audio, react strongly without improving accuracy, or let misleading media override an otherwise correct decision.

Existing MERC and Omni-model evaluations demonstrate unequal modality contributions and vulnerability to conflict~\cite{chen2025grpo,han2025camer,murzaku2025omnivox,sun2026emomm}. Most compare naturally aligned inputs or modality ablations. Ablation measures whether a channel contributes useful information; replacement asks whether unrelated but plausible in-domain content can redirect the same prediction. These questions require matched comparisons with the explicitly supplied target transcript and preceding textual context held fixed. A loss relative to correctly matched multimodal input can reflect removal of useful evidence, whereas a loss relative to modality omission establishes additional mismatch-specific harm.

We introduce a controlled reliance audit with seven conditions: Text, Text+Audio (T+A), Text+Video (T+V), Full, Random Audio (RandA), Random Video (RandV), and Random A/V (RandA/V). Random media come from another dialogue in the same dataset under label-agnostic mappings shared across models. We distinguish \textbf{utility}, the benefit of genuine target modalities; \textbf{sensitivity}, changes in predictions after replacement; and \textbf{interference}, assessed against both matched Full input and omission baselines. Evaluating three source mappings tests whether the findings depend on particular source--target pairings. A separate Qwen/MELD intervention study examines decision-query connections to mismatched audio. Extended methods and complete statistical results are provided in the appendix.

Our contributions are threefold:
\begin{enumerate}
\item A matched seven-condition audit that varies target audio and video while preserving the transcript and prior textual dialogue context.
\item Behavioral evidence separating useful-modality contribution, replacement sensitivity and mismatch-specific harm across three Omni models, two datasets and three source mappings.
\item A mechanistic confirmation in new Qwen RandA-Harm cases: decision-query Audio KO yields greater gold-vs-best-candidate margin recovery in Late than Early or Middle blocks relative to Video Sham.
\end{enumerate}

The primary evaluation finds positive T+A Accuracy changes in all six model--dataset pairs, with four significant after FDR correction. Across mappings, all 18 RandA/V-minus-Full comparisons are negative and significant under both common-valid and failed-as-incorrect policies. Effects relative to omission vary across settings. This distinction prevents modality sensitivity from being mistaken for either reliable integration or uniformly harmful mismatch.

\section{Related Work}
\paragraph{Conversational emotion recognition.}
Speaker-state, memory and graph-based architectures establish the importance of utterance evidence and dialogue structure~\cite{ghosal2019dialoguegcn,hazarika2018conversational,hu2021mmgcn,majumder2019dialoguernn}. Later work examines modality/context disentanglement, affective representations and modality selection~\cite{li2023revisiting,seong2026emotion,chen2025grpo}. Such task-specific systems are usually evaluated on naturally aligned inputs. Their ablations quantify useful contributions but do not directly measure how unrelated media changes the same decision with transcript and context fixed.

\paragraph{Omni models and modality conflict.}
OmniVox evaluates Omni-LLMs across modality and context configurations~\cite{murzaku2025omnivox}. CA-MER and EmoMM expose modality biases when emotion-bearing evidence conflicts or is missing~\cite{han2025camer,sun2026emomm}. Our audit complements these studies through label-agnostic, within-dataset cross-dialogue replacement and paired comparisons with both Full and omission baselines. It targets reliance on the complete audio/video channels rather than isolating acoustic prosody or estimating deployment failure rates.

\paragraph{Mechanistic multimodal analysis.}
NOTICE studies controlled cross-modal corruption, VisiPruner analyzes layer-wise interaction, and EmoMM steers attention under modality conflict~\cite{golovanevsky2025notice,fan2025visipruner,sun2026emomm}. Attention patterns can nominate intervention sites but do not themselves establish causal importance. We therefore separate exploratory localization from confirmatory decision-query audio-edge masking in new RandA-Harm cases. The resulting evidence concerns the tested intervention sites and cohort; it does not identify where the original prediction error first arises.

\section{Controlled Modality-Intervention Framework}
\subsection{Task Formulation}
For a target utterance $u_i$ with gold emotion $y_i\in\mathcal Y$, let $T_i$, $A_i$ and $V_i$ denote its transcript, audio and video. The textual context $C_i$ contains speaker information and at most ten preceding utterances; future turns and gold labels are excluded. Full input to an Omni model $f_\theta$ yields
\begin{equation}
\hat y_i^F=f_\theta(C_i,T_i,A_i,V_i).
\end{equation}
We hold $C_i$ and the explicit transcript $T_i$ fixed while varying modality availability or source content.

\subsection{Evaluation Data and Omni Models}
MELD's official test split contains 2,610 utterances and seven classes: anger, disgust, fear, joy, neutral, sadness and surprise~\cite{poria2019meld}. For IEMOCAP, we retain all 7,380 utterances labeled angry, excited, frustrated, happy, neutral or sad across 151 dialogues in Sessions~1--5~\cite{busso2008iemocap}. We neither merge these labels nor retain other target annotations.

We evaluate Qwen2.5-Omni-7B, Baichuan-Omni-1.5 and MiniCPM-o~2.6 off-the-shelf without task-specific fine-tuning~\cite{xu2025qwen,li2025baichuan,openbmb2025minicpmo}. Within each model, prompts preserve the context, transcript, label order and output format. Availability instructions adapt to the supplied modalities; Full and all replacement conditions share the same audiovisual instructions. Decoding and frame-sampling settings remain fixed within each model. All runs use batch size~1 and at most 160 generated tokens. Qwen and Baichuan use greedy decoding; MiniCPM uses deterministic decoding with internal beam size~3.

For Baichuan and MiniCPM on MELD, every video-bearing condition uses a common silent-MP4 re-encoding procedure with separate mono 16\,kHz WAV audio. Prepared-media hashes are checked across conditions. MiniCPM retains its duration-based A/V packing, so audio duration can change the supplied frame count. Baichuan processes decoded visual frames without converting embedded MP4 audio into audio tokens. Appendix~H documents checkpoints, prompts, media handling and available runtime records.

\subsection{Seven Matched Input Conditions}
Table~\ref{tab:conditions} defines all conditions. A fixed mapping $\pi(i)$ selects the unrelated source: RandA substitutes $A_{\pi(i)}$, RandV substitutes $V_{\pi(i)}$, and RandA/V substitutes both. Full is the matched-input reference; Text is the baseline for genuine-modality utility.

\begin{table}[H]
\caption{Seven matched input conditions. Text denotes the fixed target transcript plus preceding textual dialogue context.}\label{tab:conditions}
\centering\tighttable
\begin{adjustbox}{max width=\linewidth}
\begin{tabular}{lccc}\toprule
\textbf{Condition}&\textbf{Text/Context}&\textbf{Audio}&\textbf{Video}\\\midrule
Text & Yes & -- & --\\
T+A & Yes & Correct & --\\
T+V & Yes & -- & Correct\\
Full & Yes & Correct & Correct\\
RandA & Yes & $A_{\pi(i)}$ & Correct\\
RandV & Yes & Correct & $V_{\pi(i)}$\\
RandA/V & Yes & $A_{\pi(i)}$ & $V_{\pi(i)}$\\\bottomrule
\end{tabular}
\end{adjustbox}
\end{table}

\subsection{Random-Modality Construction and Leakage Control}
Mappings are generated before inference, fixed throughout execution and shared across models and replacement conditions. Sources must come from another dialogue in the same dataset; IEMOCAP sources must additionally share the target session. Source selection is label-agnostic and may include annotations outside the six IEMOCAP target classes. The primary mapping uses seed~42; seeds~43 and~44 provide independent additional mappings under the same restrictions. RandA/V always draws audio and video from the same source utterance, preserving their mutual correspondence while breaking alignment with the target transcript/context.

This is an in-domain stress test, not synthetic noise or an estimate of real-world mismatch frequency. Complete-speech replacement changes lexical content available through audio as well as paralinguistic cues. The explicit transcript stays fixed, but the intervention does not isolate prosody.

\subsection{Utility, Sensitivity, and Interference}
We report Accuracy as the primary metric and Weighted F1 as an additional metric, supplemented by Macro-F1 and Balanced Accuracy. Genuine-modality utility compares T+A, T+V or Full with Text. Sensitivity is the Prediction Change Rate (PCR) between Full and replacement condition $R$:
\begin{equation}
\mathrm{PCR}(R)=\frac{1}{|\mathcal V_R|}\sum_{i\in\mathcal V_R}\ind[\hat y_i^F\ne\hat y_i^R],
\end{equation}
where $\mathcal V_R$ contains targets with valid outputs in both conditions and $\ind$ is the indicator function. PCR counts any decision change, regardless of correctness.

\begin{figure}[H]
\centering\includegraphics[width=.90\linewidth,trim=140 125 140 85,clip]{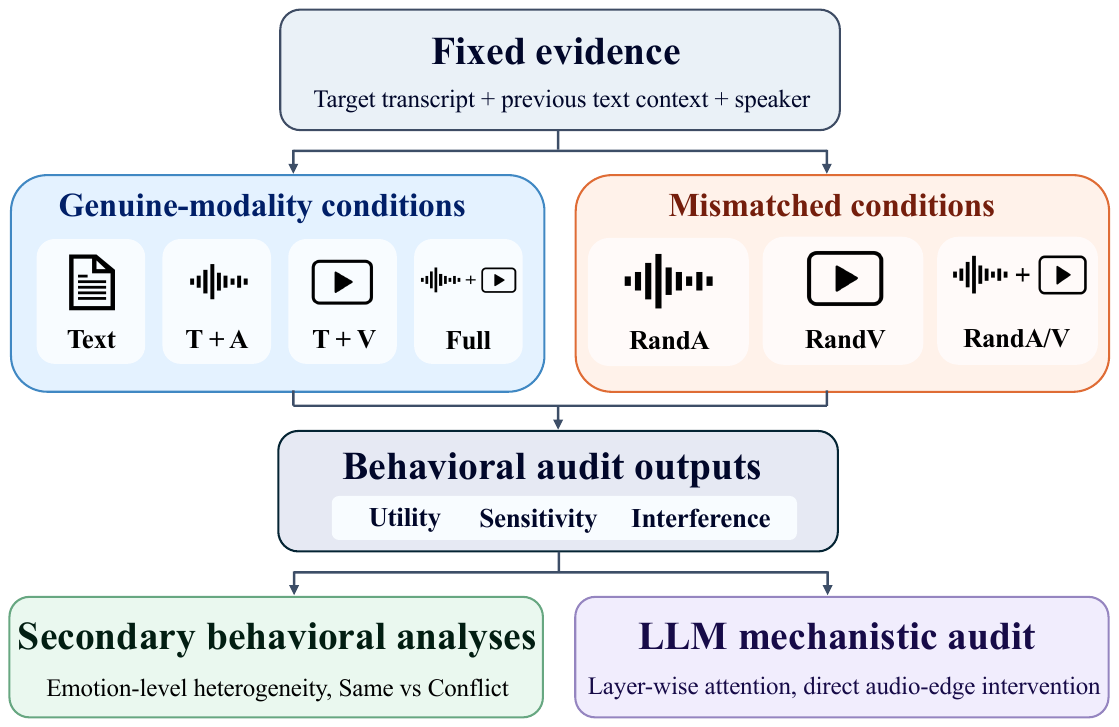}
\caption{Controlled reliance audit. Target transcript, prior textual context and speaker information remain fixed. Genuine inputs measure utility; replacements measure sensitivity and effects relative to Full or omission baselines. Emotion-level and mechanistic analyses provide follow-up diagnostics.}\label{fig:overview}
\end{figure}

Interference has two baselines. Rand-minus-Full measures a \textbf{replacement penalty}. RandA-minus-T+V, RandV-minus-T+A and RandA/V-minus-Text test \textbf{mismatch-specific harm beyond omission}. A modality can be sensitive to replacement without producing negative effects relative to omission. Figure~\ref{fig:overview} summarizes the audit. Appendix~A gives full Benefit/Harm definitions and denominators.

\subsection{Secondary Analyses and Mechanistic Case Study}
Exploratory analyses examine emotion-level heterogeneity and source-label compatibility after predictions are fixed. Same sources share the target gold label; Conflict sources do not. Both are cross-dialogue replacements. Conditional Harm is the fraction of Full-correct targets that become incorrect after replacement. Source-conflict Accuracy retains all source-annotated targets, scoring failed outputs as incorrect; conditional Harm uses common-valid Full/replacement pairs. Valence summaries retain all Full-correct targets and count failed replacement outputs as incorrect. Label-standardized Accuracy and conditional Harm assess associations with conflict, rather than identifying a causal emotion mechanism. IEMOCAP sources annotated exactly \texttt{xxx} after normalization (the undefined source annotation) are excluded only from these compatibility analyses. Full definitions and exploratory methods appear in Appendices~A--B.

The mechanistic study tests Qwen/MELD RandA-Harm cases, correct under Full but incorrect under RandA. Stable-Correct cases serve as a control. A discovery set of 50 gold-label-matched samples per group nominates fixed blocks. The new confirmatory RandA-Harm cohort is dialogue-disjoint from RandA-Harm discovery: 112 nominal cases yield 106 causal-eligible samples from 76 dialogues after prespecified parity and sham-feasibility checks.

Audio knockout (Audio KO) masks all attention logits from the single decode query predicting the first emotion token after \texttt{Emotion:} to audio-key positions, before softmax, in Early L0--8, Middle L9--18 or Late L19--27. L19--23 is a secondary candidate subset. Prefill activations are unchanged. Video Sham matches block span and masked-edge count, but not attention mass or functional importance.

Teacher-forced candidate-continuation log-likelihood $\ell_i(y)$ defines the primary margin
\begin{equation}
m_i=\ell_i(y_i)-\max_{y\ne y_i}\ell_i(y).
\end{equation}
This is a gold-vs-best-candidate log-likelihood margin, not a normalized class probability. The primary intervention outcome is the Audio-KO-minus-Video-Sham change in $m_i$; a positive effect favors the gold label. Appendices~A and~G detail scoring, cohort checks and control analyses.

\subsection{Statistical Testing and Reproducibility}
Utterances share speakers and context within dialogues. We therefore retain utterance-weighted paired Accuracy effects while resampling and sign-flipping entire dialogues. Primary seed-42 CIs and $p$ values use 1,000 dialogue-cluster bootstrap and sign-flip resamples. BH-FDR is applied separately to 18 utility comparisons, 18 omission comparisons and six RandA/V-minus-Full comparisons. Throughout, $q$ denotes the adjusted $p$ value within the specified family.

The three-mapping analysis uses 10,000 resamples and separate 18-test replacement and 54-test omission BH families under each failed-output policy. Primary failed-output sensitivity uses 10,000 resamples with separate 18-test utility, 18-test omission and six-test replacement families. Repeated targets across mappings are not pooled as independent observations; mapping means/ranges are descriptive. Full effects, CIs and family definitions are provided in Appendices~C, I--K.

Aggregate Accuracy, Weighted F1, Macro-F1 and Balanced Accuracy are computed over all expected targets. Invalid/OOM outcomes contribute no true positives and remain in gold-class support. Primary paired effects and PCR use common-valid pairs. Consequently, aggregate differences and paired effects can differ. We also test all-target paired effects with failures scored as incorrect. Final outcomes, exclusions and unchanged-input OOM retries are retained, not silently dropped (Appendix~K).

Exploratory source-conflict analyses standardize target-label distributions. Broad comparisons use 1,000 bootstrap/within-label permutation resamples and endpoint-specific 18-test BH families; three focused follow-ups additionally use dialogue-cluster bootstrap and permutations within dialogue/target-label strata, with endpoint-specific three-test BH families.

Mechanistic margin CIs use 10,000 dialogue-cluster bootstrap resamples. The prespecified tests of Late versus zero, Late minus Early and Late minus Middle use 100,000 cluster sign flips and Holm correction. Secondary block, rescue and control comparisons use their stated BH families. The accompanying materials include code, complete prompts, six mapping manifests, per-target predictions, runtime records and invalid/OOM logs.

\section{Results}
\subsection{Genuine Modalities Provide Unequal Utility}
The primary evaluation uses source-mapping seed~42. Table~\ref{tab:main} reports all seven conditions, and Table~\ref{tab:paired_summary} reports key common-valid paired effects with their separate FDR families. T+A increases aggregate Accuracy in all six model--dataset pairs; four paired gains remain significant after FDR correction. For Baichuan/MELD, the paired audio gain is +4.10\pp (95\% CI [2.38, 5.70], $q=.0045$); for MiniCPM/IEMOCAP it is +3.55\pp ([2.51, 4.59], $q=.0045$). The two remaining gains are positive but not FDR-significant.

\begin{table}[H]
\caption{Primary behavioral results (seed 42; \%). Acc.=Accuracy; WF1=Weighted F1; RandA=Random Audio; RandV=Random Video; RandA/V=Random A/V.}\label{tab:main}
\centering\tinytab
\panel{(a) MELD}
\begin{adjustbox}{max width=\linewidth}
\begin{tabular}{llccccccc}\toprule
Model&Metric&Text&T+A&T+V&Full&RandA&RandV&RandA/V\\\midrule
Qwen&Acc.&61.88&63.22&61.00&62.84&59.00&62.34&58.16\\
Qwen&WF1&60.22&61.10&58.39&60.17&55.41&59.23&54.02\\
Baichuan&Acc.&36.63&40.73&36.90&36.63&34.71&33.14&32.95\\
Baichuan&WF1&35.60&39.89&35.95&34.96&33.65&30.91&31.80\\
MiniCPM&Acc.&43.10&43.41&41.88&44.48&43.30&44.56&41.61\\
MiniCPM&WF1&43.37&43.45&42.10&44.70&43.69&44.83&41.64\\
\bottomrule\end{tabular}
\end{adjustbox}
\vspace{7pt}
\panel{(b) IEMOCAP}
\begin{adjustbox}{max width=\linewidth}
\begin{tabular}{llccccccc}\toprule
Model&Metric&Text&T+A&T+V&Full&RandA&RandV&RandA/V\\\midrule
Qwen&Acc.&49.82&51.41&48.21&50.42&46.59&49.66&45.42\\
Qwen&WF1&47.16&48.98&45.54&47.95&43.49&47.35&42.42\\
Baichuan&Acc.&46.34&50.46&45.53&49.20&45.11&47.21&43.02\\
Baichuan&WF1&45.20&49.79&44.04&48.02&43.74&46.20&41.71\\
MiniCPM&Acc.&45.07&48.63&46.02&47.76&46.78&47.43&46.04\\
MiniCPM&WF1&42.97&46.72&43.96&45.49&44.38&45.09&43.92\\
\bottomrule\end{tabular}
\end{adjustbox}
\end{table}

Additional modalities are not monotonically beneficial. Baichuan/MELD rises from 36.63\% with Text to 40.73\% with T+A, while Full returns to 36.63\%. T+A exceeds Full in five of six pairs, whereas T+V exceeds Text in only two. Utility also depends on the metric: Qwen/MELD T+A improves Accuracy and Weighted F1, but Macro-F1 falls from .462 to .460 and Balanced Accuracy from .440 to .424. Better overall Accuracy therefore does not imply uniformly better emotion-class performance. Appendices~D and~I provide all metrics and paired comparisons.

\subsection{Sensitivity, Utility, and Mismatch Vulnerability Diverge}
\begin{figure}[H]
\centering\includegraphics[width=\linewidth]{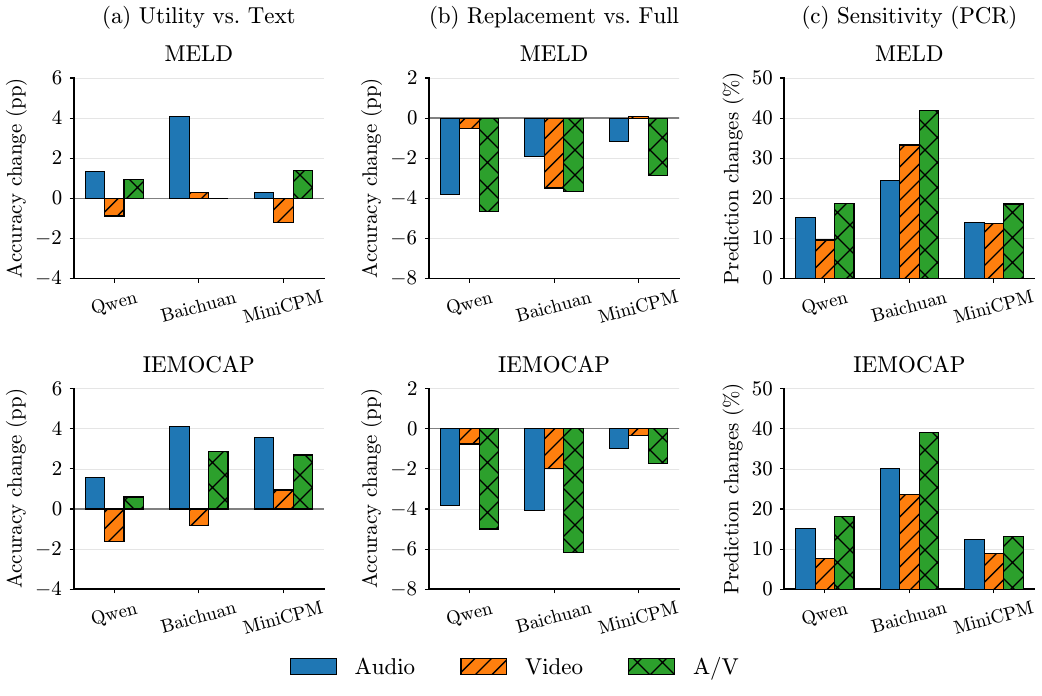}
\caption{Behavioral separation using primary seed-42 predictions. Blue, orange diagonally hatched and green cross-hatched bars denote Audio, Video and A/V. (a) Accuracy change versus Text. (b) Replacement penalty versus Full. (c) PCR on common-valid pairs. Panels (a)--(b) use aggregate Accuracy differences; Table~\ref{tab:paired_summary} gives paired effects and omission comparisons.}\label{fig:behavior}
\end{figure}

Figure~\ref{fig:behavior} contrasts genuine utility, replacement penalties and PCR. Modest net gains can coexist with substantial sensitivity. For Qwen, Full-minus-Text paired effects are +1.00\pp on MELD (95\% CI [$-0.41$, 2.39], cluster $p=.189$) and +0.59\pp on IEMOCAP ([$-0.58$, 1.76], $p=.356$). Yet RandA changes 15.12\% of common-valid MELD predictions, and RandA/V changes 18.78\%. Baichuan/MELD RandA/V changes 41.80\%. PCR also counts wrong-to-right and wrong-to-different-wrong transitions, so it can greatly exceed the net Accuracy difference.

RandA/V has a significant negative paired effect relative to Full in all six primary comparisons (Table~\ref{tab:paired_summary}). This does not establish that every replacement is harmful: MiniCPM/MELD RandV attains 44.56\% Accuracy versus 44.48\% for Full. The omission comparisons distinguish replacement penalties from additional mismatch-specific harm. Baichuan/MELD RandV is 7.55\pp below T+A on common-valid pairs ($q=.0022$), and Qwen/MELD RandA/V is 3.72\pp below Text ($q=.0022$).

\begin{samepage}
Conversely, MiniCPM/IEMOCAP RandA/V exceeds Text by +1.56\pp ($q=.0054$), despite a negative RandA/V-minus-Full effect. This positive omission-relative finding is sensitive to failed-output handling. A loss relative to Full therefore need not imply additional harm beyond simply omitting the replaced modalities.
\par
\end{samepage}

Together, these results show that a modality can provide little net benefit when correctly matched to the target yet still substantially influence predictions when its content is replaced. Utility and vulnerability to misleading input are therefore distinct properties of multimodal behavior. Appendix~B, Table~S2 summarizes Full--Text, RandA--Full, and RandA/V--Full aggregate Accuracy differences and PCR values for RandA and RandA/V.

\begin{table}[!htbp]
\caption{Primary common-valid paired Accuracy effects (pp), 95\% dialogue-cluster CIs and BH-FDR-adjusted $q$ values. $N$ is the number of paired valid targets. The utility and omission families include all three relevant conditions; panels show the key audio and joint-replacement contrasts. Positive omission effects are not harm.}\label{tab:paired_summary}
\centering\tinytab
\panel{(a) T+A--Text: genuine audio utility (18-test family)}
\begin{tabular}{llrrrr}\toprule
Dataset&Model&Effect&95\% CI&$q$&$N$\\\midrule
\multirow{3}{*}{MELD}&Qwen&+1.30&[-0.14, +2.66]&.1169&2606\\
&Baichuan&+4.10&[+2.38, +5.70]&.0045&2610\\
&MiniCPM&+0.27&[-1.16, +1.59]&.8177&2603\\
\midrule
\multirow{3}{*}{IEMOCAP}&Qwen&+1.65&[+0.56, +2.68]&.0072&7335\\
&Baichuan&+4.15&[+2.19, +6.18]&.0045&7373\\
&MiniCPM&+3.55&[+2.51, +4.59]&.0045&7370\\
\bottomrule\end{tabular}
\panel{(b) RandA/V--Full: replacement penalty (six-test family)}
\begin{tabular}{llrrrr}\toprule
Dataset&Model&Effect&95\% CI&$q$&$N$\\\midrule
\multirow{3}{*}{MELD}&Qwen&-4.69&[-6.16, -3.29]&.0012&2604\\
&Baichuan&-3.72&[-5.70, -1.80]&.0012&2605\\
&MiniCPM&-2.81&[-4.12, -1.46]&.0012&2598\\
\midrule
\multirow{3}{*}{IEMOCAP}&Qwen&-5.00&[-6.02, -3.98]&.0012&7344\\
&Baichuan&-6.21&[-7.97, -4.32]&.0012&7365\\
&MiniCPM&-1.24&[-2.07, -0.49]&.0070&7260\\
\bottomrule\end{tabular}
\panel{(c) RandA/V--Text: mismatch versus omission (18-test family)}
\begin{tabular}{llrrrr}\toprule
Dataset&Model&Effect&95\% CI&$q$&$N$\\\midrule
\multirow{3}{*}{MELD}&Qwen&-3.72&[-5.00, -2.33]&.0022&2605\\
&Baichuan&-3.68&[-5.57, -1.91]&.0022&2608\\
&MiniCPM&-1.39&[-2.91, +0.04]&.0820&2595\\
\midrule
\multirow{3}{*}{IEMOCAP}&Qwen&-4.39&[-5.69, -3.07]&.0022&7340\\
&Baichuan&-3.27&[-5.42, -1.26]&.0054&7373\\
&MiniCPM&+1.56&[+0.65, +2.49]&.0054&7261\\
\bottomrule\end{tabular}
\end{table}

\subsection{Exploratory Emotion and Source-Conflict Analyses}
Emotion-level effects do not follow a universal valence pattern. Class-balanced negative-class Full Accuracy is lower than positive-class Accuracy for all models on MELD but higher on IEMOCAP. Negative targets nevertheless show greater RandA PCR and conditional Harm in five of six pairs. We treat these as descriptive heterogeneity rather than a general claim that one valence is harder (Appendices~B and~F).

Source-label conflict is associated with additional harm in several exploratory settings after target-label standardization. For Qwen/MELD RandA, Conflict-minus-Same adjusted Accuracy is $-7.59$\pp and conditional Harm is +8.26\pp, with focused dialogue-cluster $q=.0170$ and $.0030$, respectively. This setting, Qwen/MELD RandA/V and Baichuan/IEMOCAP RandA/V remain significant for both endpoints in the broader 18-comparison families. These post-hoc label associations do not isolate a causal emotion mechanism. Appendices~B and~E report the focused and complete analyses.

\subsection{Mechanistic Case Study: RandA in Qwen}\label{sec:mechanistic}
\begin{figure}[H]
\centering\includegraphics[width=\linewidth]{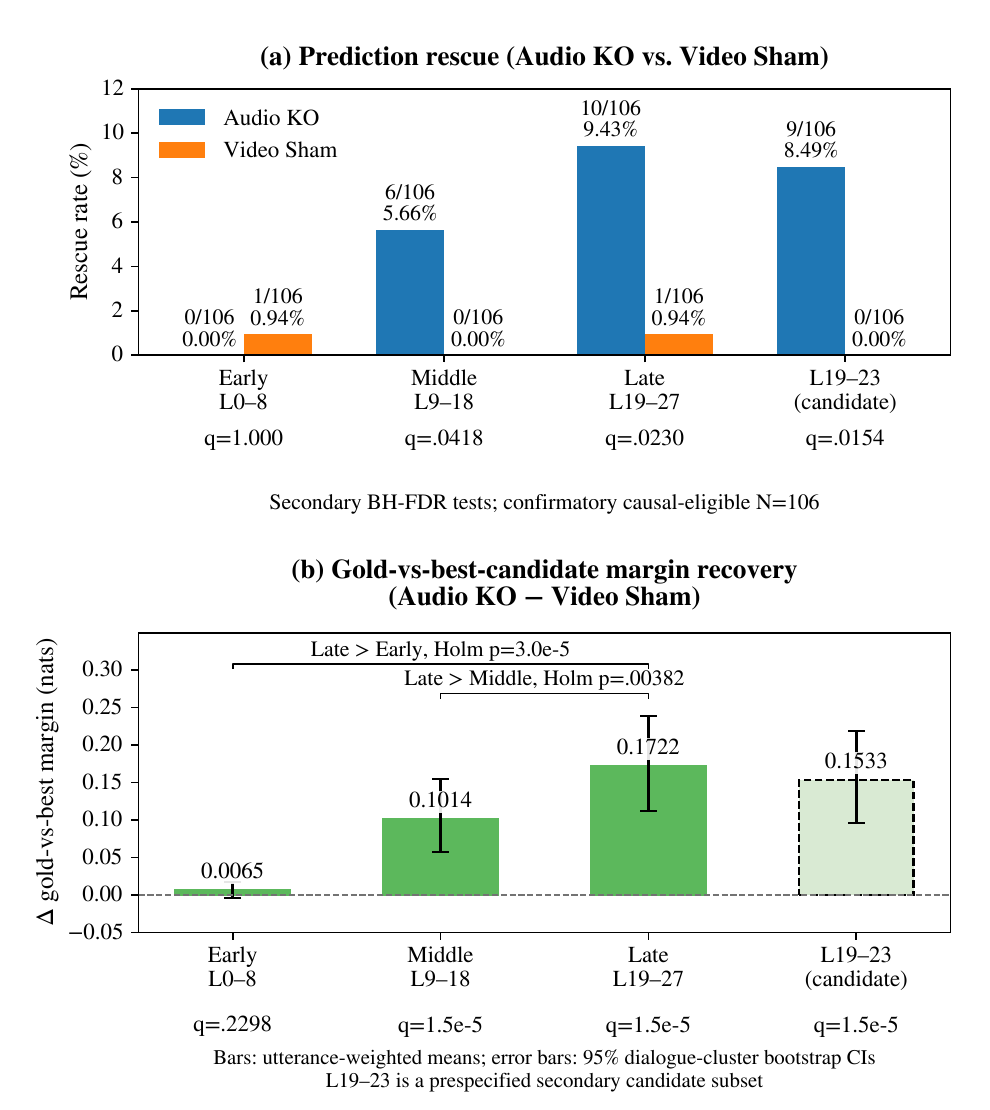}
\caption{Qwen/MELD RandA-Harm confirmation ($N=106$, 76 dialogues). (a) Prediction rescue under Audio KO and Video Sham; $q$ values are secondary BH-FDR tests. (b) Utterance-weighted Audio-KO-minus-Video-Sham margin recovery with 95\% dialogue-cluster bootstrap CIs. Brackets show prespecified Holm-corrected contrasts; L19--23 is a secondary candidate subset.}\label{fig:mechanistic}
\end{figure}

Attention patterns nominate candidate blocks. However, attention differences between RandA-Harm and Stable-Correct do not survive multiple-testing correction. Direct intervention supplies the confirmatory evidence. Figure~\ref{fig:mechanistic} reports the causal-eligible confirmatory cohort ($N=106$, 76 dialogues). Audio-KO-minus-Video-Sham margin recovery is near zero in Early, increases in Middle and is largest in Late: +0.1722 nats (95\% CI [0.1120, 0.2393]). All three prespecified contrasts pass Holm correction, including Late versus Middle ($p_{\mathrm{Holm}}=.00382$).

\begin{samepage}
Prediction rescue is secondary: negligible in Early but significant in Middle, Late and the candidate subset (Late: 9.43\%). Later Stable-Correct margin shifts are positive, but RandA-Harm recovery is significantly greater in Middle and Late under the tested masking controls. The intervention recovers some failures without locating their origin; full scores and controls are in Appendix~G.
\par
\end{samepage}

\subsection{Robustness Across Mappings and Failed-Output Policies}\label{sec:mapping_robustness}
Across seeds~42, 43 and~44, all 18 model--dataset--mapping RandA/V-minus-Full comparisons are negative and FDR-significant under common-valid dialogue-cluster inference and when invalid/OOM outcomes are retained as incorrect. These are separate mapping-specific tests, not pooled repeated targets (Appendix~J).

In the common-valid 54-test omission family, 32 comparisons have significant negative effects and seven significant positive effects. MiniCPM/IEMOCAP RandA/V-minus-Text is positive across all mappings (+1.19 to +1.62\pp), but only seed~43 remains significant when failures are scored as incorrect ($q=.04976$). Replacement penalties are robust across the tested mappings; omission-relative interpretations can depend on both model and failed-output policy. Appendix~K gives all failure counts and policy comparisons.

\section{Discussion}
\paragraph{Evaluating reliance.}
The audit distinguishes useful contribution, replacement sensitivity and harm beyond omission. Their joint reporting matters because Full is not uniformly better than T+A and omission-relative effects can differ in sign. For human--AI interaction, a model that benefits from genuine audio may still change a correct decision under unrelated input. Matched-input accuracy would miss this vulnerability, while PCR alone would conflate harmful changes with corrections. Reliability assessments should therefore include both matched-input gains and replacement-versus-omission effects. These measurements motivate future consistency checks and modality-selection strategies; testing their effectiveness and estimating deployment failure rates require separate studies.

\paragraph{Interpreting the intervention.}
The confirmation on new RandA-Harm cases shows greater decision-query margin recovery in Late than Early or Middle. Margin recovery is significantly greater for RandA-Harm than for Stable-Correct in Middle and Late under the tested masking controls. These findings concern the tested intervention sites and do not establish where the original error originates or generalize the internal dynamics beyond Qwen/MELD.

\paragraph{Limitations.}
The audit tests target-level A/V interventions in MERC on two datasets and three open-source Omni models; broader tasks, architectures and commercial systems remain future work. Three mappings establish robustness over the tested pairings rather than precisely estimating all mapping variability. Complete-speech replacement changes lexical and paralinguistic information. Source-label conflict is post-hoc and label-based, and Video Sham controls edge count without matching attention mass or functional importance. These qualifications bound the scope of the conclusions.

\clearpage
\section{Conclusion}
We presented a matched modality-intervention audit separating utility, sensitivity and mismatch-specific interference in conversational emotion recognition. Genuine audio improves Accuracy in all six model--dataset pairs, with four FDR-significant gains, while video effects are less consistent. RandA/V lowers Accuracy relative to Full in all 18 mapping comparisons under both failed-output policies, whereas omission-baseline effects vary. A Qwen/MELD case study confirms greater Late decision-query audio-edge margin recovery than Early or Middle relative to Video Sham. Multimodal evaluation should jointly measure useful contribution and behavior under misleading input.

\paragraph{Data and supplementary material.}
The appendix contains the extended methods and complete statistical tables. Reproducibility records are available at \url{https://doi.org/10.5281/zenodo.23970406}. The main results and their primary testing families are reported above.

\begin{credits}
\subsubsection{\ackname}
This research was supported by Institute of Information \& communications Technology Planning \& Evaluation (IITP) grant funded by the Korea government(MSIT) under the Artificial Intelligence Convergence Innovation Human Resources Development (IITP-2026-RS-2023-00254592) , and by ANCHOR through the Seoul ANCHOR Center, funded by the Ministry of Education (MOE) and the Seoul Metropolitan Government (2026-ANCHOR-01-007-05)

\subsubsection{\discintname}
The authors have no competing interests to declare that are relevant to the content of this article.
\end{credits}

\FloatBarrier

\clearpage
\appendix
\setcounter{table}{0}
\setcounter{figure}{0}
\setcounter{equation}{0}
\renewcommand{\thetable}{S\arabic{table}}
\renewcommand{\thefigure}{S\arabic{figure}}
\renewcommand{\theequation}{S\arabic{equation}}
\renewcommand{\theHsection}{appendix.\Alph{section}}
\renewcommand{\theHtable}{appendix.\arabic{table}}
\renewcommand{\theHfigure}{appendix.\arabic{figure}}
\renewcommand{\theHequation}{appendix.\arabic{equation}}
\section*{Appendix}
This appendix uses the task notation of the main text and provides the extended definitions, exploratory analyses, complete statistics and runtime/failure documentation. The primary seed-42 and additional-mapping results use the same reference predictions. The accompanying machine-readable files distinguish the 1,000-resample primary families from the 10,000-resample mapping and failed-output families; their adjusted $p$ values are not interchangeable.

\section{Extended Definitions and Analysis Methods}\label{sm:extended}
\subsection{Definitions and Sample-Level Outcomes}
We report Accuracy as the primary metric and Weighted F1 as an additional metric for comparability with prior MERC evaluations. Aggregate metrics are supplemented with paired sample-level measurements. For a genuine multimodal condition $M\in\{TA,TV,F\}$, Benefit and Harm relative to Text are
\begin{align}
B_i^M&=\ind[\hat y_i^T\ne y_i\wedge \hat y_i^M=y_i],\\
H_i^M&=\ind[\hat y_i^T=y_i\wedge \hat y_i^M\ne y_i].
\end{align}
Benefit measures correction of a text-only error, whereas Harm measures cases in which added target audio/video input overrides an already correct textual decision. When aggregated, Benefit is computed over Text-incorrect samples and Harm over Text-correct samples, so the two rates have different denominators. Benefit and Harm are computed only over samples with valid predictions under both Text and the corresponding multimodal condition.
We operationalize sensitivity using Prediction Change Rate (PCR) between Full and a random-modality condition $R$:
\begin{equation}
\mathrm{PCR}(R)=\frac{1}{|\mathcal{V}_R|}\sum_{i\in\mathcal{V}_R}\ind[\hat y_i^F\ne\hat y_i^R],
\end{equation}
where $\mathcal{V}_R$ contains samples with valid predictions under both conditions. PCR measures whether the decision changes, irrespective of whether the change is correct. The indicator function $\ind[E]$ equals 1 if event $E$ holds and 0 otherwise. Figure~\ref{fig:overview} summarizes the overall controlled modality-intervention framework and the follow-up analyses.

We assess interference at two complementary levels. Rand-versus-Full comparisons quantify the replacement penalty: the performance change when matched evidence is replaced by unrelated in-domain media. The stricter mismatch-specific interference comparison asks whether adding unrelated media is more harmful than omitting that modality altogether: RandA is compared with T+V, RandV with T+A, and RandA/V with Text. A negative mismatch-specific effect therefore indicates active harm beyond modality absence. This distinction is central to the audit: a model can be highly sensitive to replacement while showing either weak, strong, or even reversed active interference relative to omission. Table~\ref{tab:concepts} summarizes these operational distinctions.

\begin{table}[H]
\caption{Operational distinction among utility, sensitivity, and interference. Negative interference effects indicate harm relative to the stated baseline.}\label{tab:concepts}
\centering\small
\setlength{\tabcolsep}{5pt}\renewcommand{\arraystretch}{1.25}
\begin{adjustbox}{max width=\linewidth}
\begin{tabular}{>{\raggedright\arraybackslash}m{.19\linewidth}>{\centering\arraybackslash}m{.30\linewidth}>{\raggedright\arraybackslash}m{.40\linewidth}}
\toprule
\textbf{Concept}&\textbf{Primary metric}&\textbf{Interpretation}\\\midrule
Utility&$\Delta$Acc vs. Text&Benefit from genuine A/V\\[4pt]
Sensitivity&PCR&Prediction changes after replacement\\[4pt]
\multirow{2}{*}{Interference}&$\Delta$Acc: Rand--Full&Replacement penalty\\[-2.75pt]
&$\Delta$Acc: Rand--omission&Mismatch-specific harm\\
\bottomrule
\end{tabular}
\end{adjustbox}
\end{table}

\subsection{Exploratory and Mechanistic Methods}
We use two secondary behavioral analyses to characterize how aggregate modality effects vary across emotions and to examine what factors are associated with greater mismatch harm. First, we examine emotion-level heterogeneity to determine whether the aggregate modality effects are broadly shared across classes or concentrated in particular emotions. For MELD, Positive contains joy and Negative contains anger, disgust, fear, and sadness; for IEMOCAP, Positive contains happy and excited, and Negative contains angry, frustrated, and sad. Neutral classes, and MELD surprise, are excluded from the valence comparison. Because class frequencies differ, accuracy is macro-averaged across emotion classes within each group. This analysis is used to characterize how modality effects vary across emotions and broad valence groups, rather than to test whether positive or negative emotions are generally more difficult to recognize.

Second, we test source-label compatibility to distinguish harm caused by cross-dialogue replacement itself from additional harm associated with emotion-label conflict. After all predictions are fixed, a random source is labeled Same when its dataset gold emotion matches the target gold emotion and Conflict otherwise. For IEMOCAP, random-source cases carrying the undefined source annotation \texttt{xxx} after normalization are excluded from the source-label compatibility analysis only. All other source labels are retained; labels that do not match the target gold label are treated as Conflict. These exclusions do not alter the behavioral random-replacement analyses. This is a label-level compatibility test: the source label does not guarantee that every acoustic or visual cue in the raw media expresses that emotion. Both groups still use media from another dialogue. To reduce target-label imbalance, group accuracy is standardized to the pooled target-emotion distribution. Conditional Harm is the proportion of samples that are correct under Full but become incorrect under the mismatch condition within each compatibility group. Because this analysis is post-hoc, the three comparisons highlighted in the main text are exploratory follow-ups. They survive the focused dialogue-cluster reanalysis and remain significant for both endpoints in the broader label-adjusted 18-comparison sensitivity analysis; all 18 results are reported in Section~\ref{app:conflict}.

We additionally conduct a mechanistic case study on Qwen2.5-Omni-7B using MELD to test whether the behavioral effect of misleading audio is reflected in a targeted internal intervention. We use Qwen/MELD as a focused mechanistic case study because the behavioral analysis yields a well-defined cohort of RandA-induced failure cases suitable for targeted intervention. RandA-Harm samples are correct under Full but incorrect under RandA, whereas Stable-Correct samples remain correct. An initial discovery set of 50 samples per group, matched by gold emotion, is used only to nominate fixed layer blocks. The primary confirmatory cohort contains 112 new RandA-Harm samples from dialogues disjoint from RandA-Harm discovery; prespecified parity and sham-feasibility checks leave 106 causal-eligible samples from 76 dialogues. The remaining new cases are used only in robustness summaries. Stable-Correct serves as a control for determining whether the intervention effect is specific to RandA-induced failures rather than a generic consequence of masking.

Layer-wise attention is used only to nominate candidate regions for the direct intervention. For RandA-Harm inputs, the audio-key positions correspond to the mismatched RandA audio supplied in the behavioral condition. Audio knockout (Audio KO) acts on the single decode query used to predict the first emotion-label token immediately after the generated \texttt{Emotion:} prefix. The tested blocks are Early (L0--8), Middle (L9--18), Late (L19--27), and the secondary L19--23 candidate subset. At the selected layers, all attention logits from this query to audio-key positions are masked before softmax; prefill activations are left unchanged. Video Sham matches the layer span and masked-edge count of Audio KO. It does not match the removed attention mass or the functional importance of the masked connections.

Candidate labels are scored by teacher-forced sequence log-likelihood. For a candidate continuation $y=(y_1,\ldots,y_T)$, let $x_i$ denote the conditioned multimodal input and $h_i$ the generated prefix through \texttt{Emotion:}. We compute
\begin{equation*}
\ell_i(y)=\sum_{t=1}^{T}\log p_\theta(y_t\mid x_i,h_i,y_{<t}),
\end{equation*}
and the primary score endpoint is
\begin{equation}
m_i=\ell_i(y_i)-\max_{y\ne y_i}\ell_i(y).
\end{equation}
Thus, $m_i$ is a gold-vs-best-candidate log-likelihood margin rather than a normalized seven-class probability margin. Under the tokenizer used for the MELD intervention analysis, all seven label continuations are single tokens, although the implementation scores the full continuation and therefore also supports multi-token candidates. A positive Audio-KO-minus-Video-Sham difference means that masking the targeted audio-to-decision edges moves this candidate-label margin toward the gold emotion relative to the matched control.

\subsection{Complete Statistical Protocols}
Utterances within a dialogue share speakers, conversational context, and recording conditions, so their predictions can be dependent. We therefore use dialogue as the resampling unit while retaining an utterance-weighted paired Accuracy effect. In the primary seed-42 behavioral analyses, confidence intervals use 1,000 dialogue-cluster bootstrap resamples and $p$ values use 1,000 dialogue-level sign-flip resamples. Genuine utility comparisons (T+A vs. Text, T+V vs. Text, Full vs. Text) form a separate 18-test BH-FDR family across three conditions, three models, and two datasets. Throughout the paper, $q$ denotes the Benjamini--Hochberg FDR-adjusted $p$ value within the stated testing family. Mismatch-specific interference comparisons (RandA vs. T+V, RandV vs. T+A, RandA/V vs. Text) form another separate 18-test BH-FDR family. For the six primary RandA/V-vs.-Full replacement-penalty robustness comparisons, BH-FDR is applied across the six model--dataset tests; all six remain significant (Section~\ref{app:replacement}). Exact McNemar tests are treated only as unclustered auxiliary analyses.

The additional-mapping analysis uses 10,000 dialogue-cluster bootstrap and sign-flip resamples. Under each failed-output policy, BH-FDR is applied separately to 18 RandA/V-minus-Full comparisons and 54 omission-baseline comparisons across three mappings, three models and two datasets. Effects are reported separately by mapping; their means and ranges are descriptive, and repeated targets across mappings are not pooled as independent observations. The failed-output sensitivity analysis of the primary predictions also uses 10,000 resamples, retaining separate families of 18 utility, 18 omission and six replacement comparisons. Reference predictions and runtime details for the three-mapping analysis are described in Section~\ref{app:mapping}.

For IEMOCAP source-conflict analyses, exact \texttt{xxx} source labels are excluded before standardization and resampling.
For the label-adjusted source-conflict analysis, target-label distributions are standardized to a pooled common distribution. The broad 18-comparison analysis uses 1,000 bootstrap and 1,000 within-label permutation resamples, with separate 18-test BH-FDR families for adjusted Accuracy and Conditional Harm. The three main follow-ups additionally use dialogue-cluster bootstrap and permutation within strata jointly defined by dialogue ID and target gold label, with BH-FDR over the three focused comparisons. Full details and all 18 results are in Sections~\ref{app:conflict} and~\ref{app:repro}.

For the mechanistic confirmation, the primary endpoint is the utterance-weighted Audio KO minus Video Sham change in gold-vs-best margin. Confidence intervals use 10,000 dialogue-cluster bootstrap resamples, and primary $p$ values use 100,000 dialogue-cluster sign-flip resamples. Three prespecified primary tests---Late vs. zero, Late minus Early, and Late minus Middle---are corrected with Holm's method. Secondary block-level, rescue, and control comparisons use BH-FDR within their stated families.

Aggregate Accuracy, Weighted F1, Macro-F1 and Balanced Accuracy are computed over all expected targets, retaining invalid outputs and out-of-memory (OOM) failures as incorrect predictions. PCR and the primary paired Accuracy estimates use only samples with valid predictions under both compared conditions. Differences of condition-wise aggregate accuracies (Table~\ref{tab:effects}) can therefore differ from common-valid paired effects (Sections~\ref{app:replacement} and~\ref{app:paired}). We additionally repeat the key paired comparisons on all expected targets, treating failed outcomes as incorrect. These policies estimate different quantities and can change effect sizes and significance. Condition-wise failures, exclusions and policy comparisons are reported in Section~\ref{app:failures}. In the post-hoc conflict analysis, standardized Accuracy treats invalid/OOM outcomes as incorrect, whereas conditional Harm excludes invalid outcomes, using common-valid Full/mismatch pairs. Decoding is deterministic but model-specific. Qwen and Baichuan use deterministic greedy decoding, whereas MiniCPM uses deterministic decoding with an internal beam size of 3. All runs use batch size 1 and a maximum generation length of 160 tokens. Model checkpoints, prompt specifications, media handling, recorded runtime versions, and unrecorded fields are detailed in Section~\ref{app:repro}. In particular, Baichuan processes video inputs as decoded visual frames only; embedded MP4 audio is not converted into audio tokens.

\clearpage
\section{Additional Behavioral Summaries}\label{sm:summaries}
\subsection{Aggregate Effects and Prediction Changes}
The following table reports condition-wise aggregate Accuracy differences and common-valid PCR. Aggregate differences need not equal paired common-valid Accuracy effects because their denominators differ.
\begin{table}[H]
\caption{Summary of multimodal effects in the primary evaluation (source-mapping seed 42). $\Delta$Accuracy is in percentage points (pp), and PCR in percent. Full--Text reflects genuine-input utility; RandA--Full and RandA/V--Full are replacement penalties relative to matched Full input. Entries are differences of condition-wise aggregate accuracies, not common-valid paired estimates. Mismatch-specific interference is analyzed in Section~\ref{app:paired}.}\label{tab:effects}
\centering\tinytab
\begin{adjustbox}{max width=\linewidth}
\begin{tabular}{llcccc@{\hspace{10pt}}c}\toprule
\multirow{2}{*}{Dataset}&\multirow{2}{*}{Model}&\multicolumn{3}{c}{$\Delta$ Accuracy (pp)}&\multicolumn{2}{c}{PCR (\%)}\\
\cmidrule(lr){3-5}\cmidrule(lr){6-7}
&&Full--Text&RandA--Full&RandA/V--Full&RandA&RandA/V\\\midrule
\multirow{3}{*}{MELD}&Qwen&0.96&-3.83&-4.67&15.12&18.78\\
&Baichuan&0.00&-1.92&-3.68&24.34&41.80\\
&MiniCPM&1.38&-1.19&-2.87&13.98&18.59\\
\midrule
\multirow{3}{*}{IEMOCAP}&Qwen&0.60&-3.83&-5.00&15.11&18.06\\
&Baichuan&2.86&-4.09&-6.18&30.18&38.98\\
&MiniCPM&2.70&-0.99&-1.72&12.48&13.13\\
\bottomrule\end{tabular}
\end{adjustbox}
\end{table}\subsection{Secondary Analysis: Emotion-Level Heterogeneity}
The effect of modality interventions varies across emotions and is not explained by a consistent positive-versus-negative pattern. Under class-balanced Full accuracy, negative classes perform worse than the positive group on MELD for all three models, whereas the direction reverses on IEMOCAP. Large differences also appear within the same valence group: for Qwen on IEMOCAP, Full Accuracy is 65.71\% for happy but only 8.93\% for excited. At the same time, negative targets show higher RandA PCR and conditional harm in five of six model--dataset pairs.

These results show that broad valence groups alone do not explain which emotions are difficult or vulnerable to modality replacement. We therefore treat the valence comparison as a secondary descriptive analysis of emotion-level variation rather than as evidence that positive or negative emotions are generally more difficult to recognize. Full valence-group results are provided in Section~\ref{app:emotion}.

\subsection{Source-Label Conflict Is Associated with Greater Mismatch Harm}
A mismatched modality may be harmful simply because it comes from another dialogue. We therefore ask whether mismatch harm is greater when the emotion label of the replacement source conflicts with the target label. This is a post-hoc, exploratory analysis rather than a prespecified confirmatory analysis. Table~\ref{tab:focused} reports the three focused comparisons that remain significant after target-label standardization, dialogue-cluster inference, and BH-FDR correction. These same three settings also remain significant for both endpoints in the broader label-adjusted 18-comparison BH families; the full analysis is reported in Section~\ref{app:conflict}.

For Qwen on MELD, conflicting RandA reduces adjusted accuracy by 7.59\pp relative to Same and increases conditional harm by 8.26\pp. The pattern persists under RandA/V. For Baichuan on IEMOCAP, conflicting RandA/V reduces adjusted Accuracy by 10.61\pp and increases conditional Harm by 12.11\pp. These results indicate that mismatch harm is not explained solely by replacing target media with content from another dialogue; in some settings, source-target label conflict is associated with greater harm.

\begin{table}[H]
\caption{Focused dialogue-cluster robustness for the three exploratory source-label conflict comparisons. Effects are Conflict--Same. Positive conditional harm means that conflicting mismatches more often break a prediction that was correct under Full.}\label{tab:focused}
\centering\tinytab
\begin{adjustbox}{max width=\linewidth}
\begin{tabular}{lllcccc}\toprule
Model&Dataset&Condition&\multicolumn{2}{c}{Adjusted Accuracy}&\multicolumn{2}{c}{Conditional Harm}\\
\cmidrule(lr){4-5}\cmidrule(lr){6-7}
&&&$\Delta$ (pp)&$q$&$\Delta$ (pp)&$q$\\\midrule
Qwen&MELD&RandA&-7.59&.0170&8.26&.0030\\
Qwen&MELD&RandA/V&-8.06&.0105&9.11&.0015\\
Baichuan&IEMOCAP&RandA/V&-10.61&.0030&12.11&.0015\\
\bottomrule\end{tabular}
\end{adjustbox}
\end{table}

\section{Main RandA/V Replacement-Penalty Robustness}\label{app:replacement}
Table~\ref{tab:replacement} reports the primary seed-42 dialogue-cluster analysis for the six comparisons between RandA/V and Full. The additional-mapping analysis is reported separately in Section~\ref{app:mapping}. Effects and confidence intervals use the same utterance-weighted estimand. BH-FDR correction is applied across the six model--dataset comparisons. The effect direction is negative in all six pairs, and all six remain statistically significant after accounting for dialogue dependence and BH-FDR correction.

\begin{table}[H]
\caption{Dialogue-cluster robustness for RandA/V replacement penalty relative to Full Accuracy. CIs use 1,000 dialogue-cluster bootstrap resamples; $p$ values use 1,000 dialogue-level sign-flip resamples; $q$ values are BH-FDR adjusted across the six comparisons.}\label{tab:replacement}
\centering\tinytab
\panel{(a) MELD}
\begin{adjustbox}{max width=\linewidth}
\begin{tabular}{lcccc}\toprule
Model&$\Delta$Acc (pp)&95\% CI&$q$&$N$ / Dlg.\\\midrule
Qwen&-4.69&[-6.16,-3.29]&.0012&2604 / 280\\
Baichuan&-3.72&[-5.70,-1.80]&.0012&2605 / 280\\
MiniCPM&-2.81&[-4.12,-1.46]&.0012&2598 / 280\\
\bottomrule\end{tabular}
\end{adjustbox}
\vspace{7pt}
\panel{(b) IEMOCAP}
\begin{adjustbox}{max width=\linewidth}
\begin{tabular}{lcccc}\toprule
Model&$\Delta$Acc (pp)&95\% CI&$q$&$N$ / Dlg.\\\midrule
Qwen&-5.00&[-6.02,-3.98]&.0012&7344 / 151\\
Baichuan&-6.21&[-7.97,-4.32]&.0012&7365 / 151\\
MiniCPM&-1.24&[-2.07,-0.49]&.0070&7260 / 151\\
\bottomrule\end{tabular}
\end{adjustbox}
\end{table}

\clearpage
\section{Behavioral Metric Robustness and Benefit/Harm}\label{app:behavior}
Table~\ref{tab:robust} reports Macro-F1 and Balanced Accuracy from the same stored predictions used in the main behavioral audit. These metrics are robustness diagnostics rather than additional primary endpoints.

\begin{table}[H]
\caption{Macro-F1 and Balanced Accuracy in the primary evaluation (source-mapping seed 42). Both metrics use all expected targets, retaining invalid and OOM outcomes as incorrect predictions. MF1=Macro-F1; BA=Balanced Accuracy (0--1 scale).}\label{tab:robust}
\centering\tinytab
\panel{(a) MELD}
\begin{adjustbox}{max width=\linewidth}
\begin{tabular}{llccccccc}\toprule
Model&Metric&Text&T+A&T+V&Full&RandA&RandV&RandA/V\\\midrule
Qwen&MF1&.462&.460&.435&.449&.389&.447&.378\\
Qwen&BA&.440&.424&.398&.408&.356&.406&.346\\
Baichuan&MF1&.314&.352&.283&.296&.263&.269&.246\\
Baichuan&BA&.365&.418&.325&.345&.308&.319&.291\\
MiniCPM&MF1&.367&.389&.363&.389&.382&.390&.361\\
MiniCPM&BA&.427&.455&.416&.446&.436&.441&.414\\
\bottomrule\end{tabular}\end{adjustbox}\par
\vspace{10pt}
\panel{(b) IEMOCAP}
\begin{adjustbox}{max width=\linewidth}
\begin{tabular}{llccccccc}\toprule
Model&Metric&Text&T+A&T+V&Full&RandA&RandV&RandA/V\\\midrule
Qwen&MF1&.450&.471&.439&.464&.420&.458&.406\\
Qwen&BA&.475&.500&.467&.494&.450&.482&.431\\
Baichuan&MF1&.441&.489&.443&.481&.440&.460&.418\\
Baichuan&BA&.448&.512&.471&.520&.470&.491&.444\\
MiniCPM&MF1&.406&.443&.417&.432&.422&.426&.416\\
MiniCPM&BA&.436&.469&.453&.467&.457&.459&.449\\
\bottomrule\end{tabular}\end{adjustbox}\par
\end{table}

\clearpage
Table~\ref{tab:benefit} reports genuine-modality Benefit and Harm relative to Text. Benefit is the fraction of Text-incorrect samples corrected by the added genuine modality; Harm is the fraction of Text-correct samples made incorrect.

\begin{table}[H]
\caption{Genuine-modality Benefit and Harm relative to Text. Each entry shows percentage and numerator/denominator. Rates are computed over common-valid Text/condition pairs.}\label{tab:benefit}
\centering\tinytab
\panel{(a) MELD}
\begin{adjustbox}{max width=\linewidth}
\begin{tabular}{lccc}\toprule
Model&Condition&Benefit \% ($n/d$)&Harm \% ($n/d$)\\\midrule
\multirow{3}{*}{Qwen}&T+A&17.05 (169/991)&8.36 (135/1615)\\
&T+V&13.71 (136/992)&9.80 (158/1613)\\
&Full&19.56 (194/992)&10.42 (168/1613)\\
\addlinespace[3pt]
\multirow{3}{*}{Baichuan}&T+A&18.38 (304/1654)&20.61 (197/956)\\
&T+V&18.29 (302/1651)&30.71 (293/954)\\
&Full&18.63 (308/1653)&32.08 (306/954)\\
\addlinespace[3pt]
\multirow{3}{*}{MiniCPM}&T+A&9.68 (143/1478)&12.09 (136/1125)\\
&T+V&9.60 (141/1469)&15.09 (169/1120)\\
&Full&11.30 (167/1478)&11.67 (131/1123)\\
\addlinespace[3pt]
\bottomrule\end{tabular}
\end{adjustbox}
\vspace{7pt}
\panel{(b) IEMOCAP}
\begin{adjustbox}{max width=\linewidth}
\begin{tabular}{lccc}\toprule
Model&Condition&Benefit \% ($n/d$)&Harm \% ($n/d$)\\\midrule
\multirow{3}{*}{Qwen}&T+A&14.96 (548/3663)&11.63 (427/3672)\\
&T+V&10.51 (385/3663)&13.67 (502/3673)\\
&Full&15.91 (583/3665)&14.69 (540/3675)\\
\addlinespace[3pt]
\multirow{3}{*}{Baichuan}&T+A&31.83 (1259/3955)&27.88 (953/3418)\\
&T+V&29.59 (1171/3957)&35.92 (1227/3416)\\
&Full&34.12 (1350/3957)&33.21 (1134/3415)\\
\addlinespace[3pt]
\multirow{3}{*}{MiniCPM}&T+A&14.49 (586/4044)&9.74 (324/3326)\\
&T+V&11.08 (445/4016)&10.96 (363/3312)\\
&Full&13.24 (535/4040)&10.03 (333/3321)\\
\addlinespace[3pt]
\bottomrule\end{tabular}
\end{adjustbox}
\end{table}

\clearpage
\section{Full Source-Label Conflict Sensitivity Analysis}\label{app:conflict}
The broad post-hoc sensitivity analysis evaluates all 18 combinations of model, dataset, and mismatch condition after target-label standardization. For IEMOCAP, source cases labeled exactly \texttt{xxx} are excluded before Same/Conflict grouping; all other source labels are retained. To preserve readability, MELD and IEMOCAP are reported as two panels. The $q$ columns use separate 18-test BH-FDR families for adjusted Accuracy and Conditional Harm. Table~\ref{tab:conflict} reports the complete 18 label-adjusted source-conflict comparisons.

\begin{table}[H]
\caption{All 18 label-adjusted source-conflict comparisons. Effects are Conflict--Same after target-label standardization.}\label{tab:conflict}
\centering\tinytab
\panel{(a) MELD}
\begin{adjustbox}{max width=\linewidth}
\begin{tabular}{lccccccc}\toprule
\textbf{Model}&\textbf{Condition}&\multicolumn{3}{c}{Adjusted Accuracy}&\multicolumn{3}{c}{Conditional Harm}\\
\cmidrule(lr){3-5}\cmidrule(lr){6-8}
&&Effect (pp)&95\% CI&$q$&Effect (pp)&95\% CI&$q$\\\midrule
\multirow{3}{*}{Qwen}&RandA&-7.59&[-12.17,-3.26]&.0090&+8.26&[+5.21,+11.37]&.0036\\
&RandV&-2.49&[-7.05,+2.17]&.3480&+0.92&[-2.34,+3.77]&.5485\\
&RandA/V&-8.06&[-12.19,-3.80]&.0108&+9.11&[+5.39,+12.95]&.0036\\
\multirow{3}{*}{Baichuan}&RandA&-2.23&[-6.51,+1.69]&.3480&-2.44&[-9.59,+4.76]&.5204\\
&RandV&-2.96&[-6.87,+1.10]&.2479&+2.19&[-4.45,+8.99]&.5485\\
&RandA/V&-4.82&[-8.98,-1.04]&.0877&+4.08&[-3.16,+10.76]&.4177\\
\multirow{3}{*}{MiniCPM}&RandA&-5.35&[-10.06,-0.63]&.0877&+1.98&[-2.94,+6.03]&.5204\\
&RandV&-4.81&[-9.29,-0.24]&.1239&+2.07&[-1.38,+5.23]&.4855\\
&RandA/V&-4.16&[-8.66,+0.49]&.1780&+4.67&[+0.32,+8.92]&.1782\\
\bottomrule\end{tabular}
\end{adjustbox}

\vspace{7pt}
\panel{(b) IEMOCAP}
\begin{adjustbox}{max width=\linewidth}
\begin{tabular}{lccccccc}\toprule
\textbf{Model}&\textbf{Condition}&\multicolumn{3}{c}{Adjusted Accuracy}&\multicolumn{3}{c}{Conditional Harm}\\
\cmidrule(lr){3-5}\cmidrule(lr){6-8}
&&Effect (pp)&95\% CI&$q$&Effect (pp)&95\% CI&$q$\\\midrule
\multirow{3}{*}{Qwen}&RandA&-3.70&[-6.86,-0.75]&.0629&+5.76&[+2.65,+8.80]&.0051\\
&RandV&-2.22&[-5.14,+0.84]&.2434&+3.00&[+1.03,+4.91]&.0427\\
&RandA/V&-5.62&[-8.55,-2.57]&.0090&+8.03&[+5.22,+11.01]&.0036\\
\multirow{3}{*}{Baichuan}&RandA&-6.59&[-9.81,-3.27]&.0090&+8.19&[+4.03,+12.32]&.0036\\
&RandV&-2.33&[-5.57,+0.76]&.2434&+4.19&[+0.61,+7.43]&.0899\\
&RandA/V&-10.61&[-13.93,-7.40]&.0090&+12.11&[+7.65,+16.37]&.0036\\
\multirow{3}{*}{MiniCPM}&RandA&-1.56&[-4.49,+1.64]&.3480&+3.28&[+0.18,+6.19]&.0659\\
&RandV&+0.61&[-2.70,+3.75]&.7193&+1.98&[-0.61,+4.17]&.2652\\
&RandA/V&-2.46&[-5.77,+0.84]&.2191&+5.36&[+2.52,+7.73]&.0051\\
\bottomrule\end{tabular}
\end{adjustbox}
\end{table}

\clearpage
\section{Supplementary Emotion-Level Results}\label{app:emotion}
Table~\ref{tab:valence} provides the complete valence-group summary used for the descriptive analysis in Sec.~4.3. Full accuracy and Audio Utility are class-balanced within each valence group; PCR and Conditional Harm are sample-level rates within the group. PCR uses common-valid Full/RandA pairs; Conditional Harm uses all Full-correct targets, counting invalid/OOM RandA outcomes as incorrect. These values are descriptive and are not used to claim a universal positive-versus-negative difference.

\begin{table}[H]
\caption{Emotion-level descriptive analysis by valence group. Full Accuracy and Audio Utility are class-balanced within each group; PCR and Conditional Harm are sample-level rates. Audio Utility is T+A minus Text.}\label{tab:valence}
\centering\tinytab
\panel{(a) MELD}
\begin{adjustbox}{max width=\linewidth}
\begin{tabular}{llcccc}\toprule
Model&Group&\shortstack{Full Acc.\\(\%)}&\shortstack{Audio Utility\\(pp)}&\shortstack{RandA PCR\\(\%)}&\shortstack{Cond. Harm\\(\%)}\\\midrule
\multirow{2}{*}{Qwen}&Positive&41.04&+5.72&21.45&30.91\\
&Negative&29.06&-5.25&20.06&21.76\\
\addlinespace[3pt]
\multirow{2}{*}{Baichuan}&Positive&86.07&-1.74&10.75&7.80\\
&Negative&17.26&+4.20&29.49&35.17\\
\addlinespace[3pt]
\multirow{2}{*}{MiniCPM}&Positive&81.09&+2.24&7.48&4.91\\
&Negative&30.24&+4.77&16.92&14.53\\
\addlinespace[3pt]
\bottomrule\end{tabular}
\end{adjustbox}
\vspace{7pt}
\panel{(b) IEMOCAP}
\begin{adjustbox}{max width=\linewidth}
\begin{tabular}{llcccc}\toprule
Model&Group&\shortstack{Full Acc.\\(\%)}&\shortstack{Audio Utility\\(pp)}&\shortstack{RandA PCR\\(\%)}&\shortstack{Cond. Harm\\(\%)}\\\midrule
\multirow{2}{*}{Qwen}&Positive&37.32&+4.08&14.71&17.98\\
&Negative&46.52&+0.59&17.68&20.38\\
\addlinespace[3pt]
\multirow{2}{*}{Baichuan}&Positive&47.60&+0.91&28.70&22.95\\
&Negative&53.76&+8.46&30.99&27.31\\
\addlinespace[3pt]
\multirow{2}{*}{MiniCPM}&Positive&41.65&+0.47&11.42&8.94\\
&Negative&46.68&+7.69&12.25&9.59\\
\addlinespace[3pt]
\bottomrule\end{tabular}
\end{adjustbox}
\end{table}

\clearpage
\section{Direct Audio-Edge Intervention Statistics}\label{app:intervention}
The main mechanistic confirmation uses the primary RandA-Harm cohort, which excludes RandA-Harm discovery dialogues (112 nominal samples; 106 causal-eligible samples from 76 dialogues). The pooled RandA-Harm causal-eligible robustness set contains 193 samples. Tables~\ref{tab:score}--\ref{tab:stable} report the confirmatory score-level effects, prespecified primary tests, and Stable-Correct control results, respectively.

\begin{table}[H]
\caption{Confirmatory score-level direct audio-edge intervention results ($N=106$). Effects are utterance-weighted Audio KO minus Video Sham differences; CIs are 95\% dialogue-cluster bootstrap intervals and $q$ values are secondary BH-FDR adjusted.}\label{tab:score}
\centering\tighttable
\noindent\makebox[\linewidth][c]{\textbf{(a) Gold-label score recovery}}\par
\begin{adjustbox}{max width=\linewidth}
\begin{tabular}{lccc}\toprule
\textbf{Block}&\textbf{Score effect (nats)}&\textbf{95\% CI}&$q$\\\midrule
Early L0--8&0.0062&[-0.0003,+0.0131]&.0771\\
Middle L9--18&0.0690&[+0.0387,+0.1054]&1.5e-5\\
Late L19--27&0.1189&[+0.0804,+0.1623]&1.5e-5\\
L19--23 (candidate)&0.1062&[+0.0696,+0.1480]&1.5e-5\\\bottomrule
\end{tabular}
\end{adjustbox}\par
\vspace{12pt}
\noindent\makebox[\linewidth][c]{\textbf{(b) Gold-vs-best-candidate margin recovery}}\par
\begin{adjustbox}{max width=\linewidth}
\begin{tabular}{lccc}\toprule
\textbf{Block}&\textbf{Margin effect (nats)}&\textbf{95\% CI}&$q$\\\midrule
Early L0--8&0.0065&[-0.0042,+0.0178]&.2298\\
Middle L9--18&0.1014&[+0.0573,+0.1546]&1.5e-5\\
Late L19--27&0.1722&[+0.1120,+0.2393]&1.5e-5\\
L19--23 (candidate)&0.1533&[+0.0964,+0.2187]&1.5e-5\\\bottomrule
\end{tabular}
\end{adjustbox}
\end{table}

\begin{table}[H]
\caption{Prespecified primary confirmatory tests on the gold-vs-best-candidate margin. $p$ values use dialogue-cluster sign-flip tests and Holm correction.}\label{tab:primary}
\centering\tighttable
\begin{adjustbox}{max width=\linewidth}
\begin{tabular}{lccc}\toprule \textbf{Primary test}&\textbf{Effect (nats)}&Raw $p$&Holm $p$\\\midrule
Late vs. 0&+0.1722&1.0e-5&3.0e-5\\
Late $-$ Early&+0.1657&1.0e-5&3.0e-5\\
Late $-$ Middle&+0.0708&.00382&.00382\\\bottomrule
\end{tabular}
\end{adjustbox}
\end{table}

\begin{table}[H]
\caption{Stable-Correct confirmatory control. The first effect is Audio KO minus Video Sham within Stable-Correct; the second is the RandA-Harm minus Stable-Correct difference in margin recovery. Effects are in nats; $q$ values are BH-FDR adjusted.}\label{tab:stable}
\centering\tighttable
\begin{adjustbox}{max width=\linewidth}
\begin{tabular}{lccccc}\toprule
\textbf{Block}&\multicolumn{2}{c}{Stable-Correct}&\multicolumn{3}{c}{RandA-Harm $-$ Stable}\\
\cmidrule(lr){2-3}\cmidrule(lr){4-6}
&Effect&$q$&Contrast&95\% CI&$q$\\\midrule
Early&+0.0050&.2951&+0.0015&[-0.0128,+0.0162]&.8235\\
Middle&+0.0067&.5131&+0.0947&[+0.0446,+0.1506]&.00092\\
Late&+0.0441&.0396&+0.1281&[+0.0492,+0.2091]&.00414\\
L19--23 (candidate)&+0.0720&.00036&+0.0813&[+0.0065,+0.1575]&.05096\\\bottomrule
\end{tabular}
\end{adjustbox}
\end{table}

Stable-Correct confirmatory prediction harm is 0/112 for both Audio KO and Video Sham in all four blocks. A smaller positive score shift appears in later Stable-Correct blocks, but RandA-Harm shows significantly larger effects in Middle and Late. Across the expansion manifest, 3,128 of 3,184 nominal causal pairs are eligible: 52 exclusions arise from 13 canonical--master prediction mismatches propagated across four blocks, and four from one condition in which a matched Video Sham cannot be constructed. No OOM, runtime, or score errors occur.

The original confirmatory split excludes RandA-Harm discovery dialogues. As a post-hoc sensitivity check, additionally excluding dialogues represented in Stable-Correct discovery leaves 84 eligible cases from 60 dialogues. Late margin recovery is +0.1801 nats (95\% dialogue-cluster CI [0.1058, 0.2614]); all three sensitivity contrasts remain positive and significant after Holm correction across the three tests (Late versus zero: $p_{\mathrm{Holm}}=.00003$; Late minus Early: $.00012$; Late minus Middle: $.02805$). This stricter check supplements the original prespecified analysis; its full results are supplied in \texttt{strict\_discovery\_sensitivity.csv}.

\clearpage
\section{Reproducibility and Resampling Details}\label{app:repro}
Table~\ref{tab:runtime} summarizes the recorded model/runtime and media-processing setup.
\begin{table}[H]
\caption{Recorded model/runtime and media-processing setup. All runs use batch size 1 and a maximum generation length of 160 tokens.}\label{tab:runtime}
\centering\footnotesize
\begin{adjustbox}{max width=\linewidth}
\begin{tabular}{>{\raggedright\arraybackslash}m{.13\linewidth}>{\raggedright\arraybackslash}m{.31\linewidth}>{\raggedright\arraybackslash}m{.44\linewidth}}\toprule
\textbf{Model}&\textbf{Checkpoint / recorded runtime}&\textbf{Decoding and media handling}\\\midrule
Qwen&Qwen/Qwen2.5-Omni-7B;\newline PyTorch 2.11.0+cu128&Sampling disabled; 2 fps; min 4/max 768 frames; explicit mono 16 kHz WAV; video-path embedded audio disabled.\\
Baichuan&baichuan-inc/Baichuan-Omni-1d5;\newline PyTorch 2.4.0+cu124&Sampling disabled; explicit audio and video-frame paths processed separately. The video processor reads visual frames only; embedded MP4 audio is not converted to audio tokens.\\
MiniCPM&openbmb/MiniCPM-o-2\_6;\newline PyTorch 2.3.1+cu121&Sampling disabled; internal beam size 3; bf16/SDPA; video opened with audio disabled and separate mono 16 kHz audio chunks.\\\bottomrule
\end{tabular}
\end{adjustbox}
\end{table}

The system prompt is: ``You are an emotion analysis expert. Please infer emotion labels based on the given video/audio.'' The user prompt presents up to ten preceding turns in chronological order together with the target transcript, forbids future context, provides the dataset label set, and requests an \texttt{Emotion:} line followed by a brief \texttt{Reason:} line. The templates share a common structure, with modality-availability instructions adapted to the supplied inputs. Full and all replacement conditions use the same audiovisual instructions; complete templates are included in the accompanying reproducibility files. MELD label order is anger, disgust, sadness, joy, neutral, surprise, fear; IEMOCAP label order is angry, happy, sad, neutral, excited, frustrated. The Qwen/MELD behavioral evaluation uses a pinned checkpoint, Transformers 4.57.6 and SDPA attention; its configuration and hashes accompany the reproducibility records. Pinned checkpoint and runtime records also accompany the primary Baichuan and MiniCPM MELD evaluations; corresponding version records remain incomplete for the primary IEMOCAP runs.

The primary random-source construction uses seed 42; the additional analysis evaluates seeds 43 and 44 under the same cross-dialogue, label-agnostic rules, sharing each mapping across models and replacement conditions. IEMOCAP source selection remains restricted to the target session.

Reference predictions and runtime details for the three-mapping analysis are documented in Section~\ref{app:mapping}.
For Baichuan and MiniCPM on MELD, all video-bearing conditions use a common silent-MP4 re-encoding procedure, with audio supplied separately as mono 16 kHz WAV. Prepared-media hashes are verified across conditions. MiniCPM retains its duration-based audio--video packing; differences in audio duration can therefore change the supplied frame count.

Resampling families for the primary analyses are defined in Section~3.7. The primary seed-42 replacement analysis uses 1,000 dialogue-cluster bootstrap and 1,000 sign-flip resamples; the broad label-adjusted conflict analysis uses 1,000 bootstrap plus 1,000 within-label permutations; and the mechanistic confirmation uses 10,000 dialogue-cluster bootstrap resamples and 100,000 sign flips, with Holm correction for its three prespecified primary tests. The additional-mapping and failed-output sensitivity analyses use separate 10,000-resample protocols and the testing families specified in Section~3.7 and Sections~\ref{app:mapping} and~\ref{app:failures}.

The accompanying machine-readable tables contain all seed-specific effects, confidence intervals, adjusted $p$ values and failure counts. The accompanying reproducibility archive provides inference and analysis code, complete prompt templates, seed-specific replacement manifests and hashes, per-target predictions, invalid/OOM and retry records, environment metadata, and mechanistic cohort, intervention and score records. Raw MELD and IEMOCAP media, dataset transcripts, free-form model responses and model weights are not redistributed; records refer to dataset identifiers.

\clearpage
\section{Paired Utility and Mismatch-Specific Interference}\label{app:paired}
Tables~\ref{tab:utility} and~\ref{tab:mismatch} report the primary seed-42 cluster-aware paired analyses used to distinguish useful contribution from active mismatch harm. Utility and mismatch-specific interference are corrected in separate 18-test BH-FDR families. The tables use the same primary seed-42 predictions as the main behavioral results; additional mappings and failed-output analyses are reported separately in Sections~\ref{app:mapping} and~\ref{app:failures}.\par

\begin{table}[H]
\caption{Cluster-aware genuine-modality utility. Effects are condition minus Text Accuracy (pp). The $q$ column uses one 18-test BH-FDR family.}\label{tab:utility}
\centering\tinytab
\panel{(a) MELD}
\begin{adjustbox}{max width=\linewidth}
\begin{tabular}{llccccc}\toprule
Model&Comparison&$\Delta$Acc&95\% CI&$p$&$q$&$N$\\\midrule
\multirow{3}{*}{Qwen}&T+A--Text&+1.30&[-0.14,+2.66]&.0649&.1169&2606\\
&T+V--Text&-0.84&[-1.96,+0.34]&.1918&.2656&2605\\
&Full--Text&+1.00&[-0.41,+2.39]&.1888&.2656&2605\\
\addlinespace[3pt]
\multirow{3}{*}{Baichuan}&T+A--Text&+4.10&[+2.38,+5.70]&.0010&.0045&2610\\
&T+V--Text&+0.35&[-1.62,+2.47]&.7722&.8177&2605\\
&Full--Text&+0.08&[-1.92,+2.12]&.9720&.9720&2607\\
\addlinespace[3pt]
\multirow{3}{*}{MiniCPM}&T+A--Text&+0.27&[-1.16,+1.59]&.7652&.8177&2603\\
&T+V--Text&-1.08&[-2.43,+0.31]&.1349&.2207&2589\\
&Full--Text&+1.38&[-0.04,+2.75]&.0500&.0999&2601\\
\addlinespace[3pt]
\bottomrule\end{tabular}
\end{adjustbox}
\vspace{7pt}
\panel{(b) IEMOCAP}
\begin{adjustbox}{max width=\linewidth}
\begin{tabular}{llccccc}\toprule
Model&Comparison&$\Delta$Acc&95\% CI&$p$&$q$&$N$\\\midrule
\multirow{3}{*}{Qwen}&T+A--Text&+1.65&[+0.56,+2.68]&.0020&.0072&7335\\
&T+V--Text&-1.59&[-2.78,-0.39]&.0110&.0330&7336\\
&Full--Text&+0.59&[-0.58,+1.76]&.3556&.4573&7340\\
\addlinespace[3pt]
\multirow{3}{*}{Baichuan}&T+A--Text&+4.15&[+2.19,+6.18]&.0010&.0045&7373\\
&T+V--Text&-0.76&[-3.18,+1.51]&.5275&.6330&7373\\
&Full--Text&+2.93&[+0.50,+5.28]&.0170&.0437&7372\\
\addlinespace[3pt]
\multirow{3}{*}{MiniCPM}&T+A--Text&+3.55&[+2.51,+4.59]&.0010&.0045&7370\\
&T+V--Text&+1.12&[+0.13,+2.07]&.0250&.0562&7328\\
&Full--Text&+2.74&[+1.86,+3.60]&.0010&.0045&7361\\
\addlinespace[3pt]
\bottomrule\end{tabular}
\end{adjustbox}
\end{table}

\begin{table}[H]
\caption{Cluster-aware mismatch-specific interference. Effects are Random minus the corresponding omission baseline Accuracy (pp): RandA--T+V, RandV--T+A, and RandA/V--Text. Negative effects indicate active harm beyond modality omission. The $q$ column uses a separate 18-test BH-FDR family.}\label{tab:mismatch}
\centering\tinytab
\panel{(a) MELD}
\begin{adjustbox}{max width=\linewidth}
\begin{tabular}{llccccc}\toprule
Model&Comparison&$\Delta$Acc&95\% CI&$p$&$q$&$N$\\\midrule
\multirow{3}{*}{Qwen}&RandA--T+V&-2.00&[-2.99,-0.99]&.0010&.0022&2606\\
&RandV--T+A&-0.88&[-2.04,+0.16]&.1309&.1386&2605\\
&RandA/V--Text&-3.72&[-5.00,-2.33]&.0010&.0022&2605\\
\addlinespace[3pt]
\multirow{3}{*}{Baichuan}&RandA--T+V&-2.19&[-3.48,-0.89]&.0040&.0065&2605\\
&RandV--T+A&-7.55&[-9.34,-5.68]&.0010&.0022&2608\\
&RandA/V--Text&-3.68&[-5.57,-1.91]&.0010&.0022&2608\\
\addlinespace[3pt]
\multirow{3}{*}{MiniCPM}&RandA--T+V&+1.31&[+0.34,+2.37]&.0150&.0207&2591\\
&RandV--T+A&+1.23&[+0.04,+2.34]&.0480&.0599&2604\\
&RandA/V--Text&-1.39&[-2.91,+0.04]&.0729&.0820&2595\\
\addlinespace[3pt]
\bottomrule\end{tabular}
\end{adjustbox}
\vspace{7pt}
\panel{(b) IEMOCAP}
\begin{adjustbox}{max width=\linewidth}
\begin{tabular}{llccccc}\toprule
Model&Comparison&$\Delta$Acc&95\% CI&$p$&$q$&$N$\\\midrule
\multirow{3}{*}{Qwen}&RandA--T+V&-1.66&[-2.30,-1.03]&.0010&.0022&7344\\
&RandV--T+A&-1.79&[-2.52,-1.04]&.0010&.0022&7337\\
&RandA/V--Text&-4.39&[-5.69,-3.07]&.0010&.0022&7340\\
\addlinespace[3pt]
\multirow{3}{*}{Baichuan}&RandA--T+V&-0.42&[-1.38,+0.63]&.4396&.4396&7373\\
&RandV--T+A&-3.14&[-4.44,-1.77]&.0010&.0022&7351\\
&RandA/V--Text&-3.27&[-5.42,-1.26]&.0030&.0054&7373\\
\addlinespace[3pt]
\multirow{3}{*}{MiniCPM}&RandA--T+V&+0.81&[+0.04,+1.64]&.0500&.0599&7276\\
&RandV--T+A&-1.09&[-1.86,-0.28]&.0120&.0180&7354\\
&RandA/V--Text&+1.56&[+0.65,+2.49]&.0030&.0054&7261\\
\addlinespace[3pt]
\bottomrule\end{tabular}
\end{adjustbox}
\end{table}

\FloatBarrier
\clearpage
\section{Additional-Mapping Robustness}\label{app:mapping}
This section reports behavioral robustness across source-mapping seeds 42, 43 and 44. The primary seed-42 predictions are shared with the main behavioral evaluation; the reference configuration is described in Section~\ref{sec:reference_runtime}. Effects are reported separately by mapping, with dialogue-cluster inference. Under each failed-output policy, BH-FDR is applied separately to 18 replacement and 54 omission comparisons, using 10,000 bootstrap and sign-flip resamples. Repeated targets across mappings are not pooled as independent observations.

\begin{table}[H]
\caption{Three-mapping RandA/V-minus-Full Accuracy effects on common-valid pairs. Effects and confidence intervals are in pp; $q$ uses the separate 18-test BH-FDR family. Primary reference predictions are described in Section~\ref{sec:reference_runtime}.}\label{tab:mapping_replacement}
\centering\tinytab
\begin{adjustbox}{max width=\linewidth}
\begin{tabular}{lccccc}\toprule
Model&Seed&Effect&95\% CI&$q$&$N$\\\midrule
\multicolumn{6}{c}{\textbf{MELD}}\\
\midrule
\multirow{3}{*}{Qwen}&42&-4.69&[-6.09,-3.26]&.00018&2604\\
&43&-4.46&[-5.87,-3.04]&.00018&2603\\
&44&-4.76&[-6.25,-3.25]&.00018&2604\\
\addlinespace[3pt]
\multirow{3}{*}{Baichuan}&42&-3.72&[-5.70,-1.77]&.00055&2605\\
&43&-3.22&[-5.05,-1.43]&.00096&2605\\
&44&-3.15&[-4.92,-1.37]&.00090&2604\\
\addlinespace[3pt]
\multirow{3}{*}{MiniCPM}&42&-2.81&[-4.22,-1.44]&.00045&2598\\
&43&-2.27&[-3.49,-1.06]&.00045&2596\\
&44&-2.24&[-3.56,-0.93]&.00101&2595\\
\addlinespace[3pt]
\midrule
\multicolumn{6}{c}{\textbf{IEMOCAP}}\\
\midrule
\multirow{3}{*}{Qwen}&42&-5.00&[-5.97,-4.07]&.00018&7344\\
&43&-5.04&[-5.94,-4.16]&.00018&7341\\
&44&-4.81&[-5.81,-3.83]&.00018&7346\\
\addlinespace[3pt]
\multirow{3}{*}{Baichuan}&42&-6.21&[-7.91,-4.45]&.00018&7365\\
&43&-6.03&[-7.82,-4.19]&.00018&7368\\
&44&-6.11&[-7.85,-4.37]&.00018&7368\\
\addlinespace[3pt]
\multirow{3}{*}{MiniCPM}&42&-1.24&[-2.03,-0.44]&.00233&7260\\
&43&-1.17&[-1.93,-0.41]&.00260&7271\\
&44&-1.55&[-2.30,-0.78]&.00018&7289\\
\addlinespace[3pt]
\bottomrule\end{tabular}
\end{adjustbox}
\par\smallskip\footnotesize All 18 effects are negative and have $q<.05$ under both failed-output policies. CIs are unadjusted percentile-bootstrap intervals; significance is based on BH-FDR-adjusted sign-flip tests.
\end{table}

\begin{table}[H]
\caption{Omission-baseline effects across all three mappings on common-valid pairs. Min/max effects are in pp. Significant negative and positive counts use $q<.05$ in the separate 54-test BH-FDR family.}\label{tab:mapping_omission}
\centering\tinytab
\begin{adjustbox}{max width=\linewidth}
\begin{tabular}{llcccc}\toprule
Model&Comparison&Min.&Max.&Sig. neg.&Sig. pos.\\\midrule
\multicolumn{6}{c}{\textbf{MELD}}\\
\midrule
\multirow{3}{*}{Qwen}&RandA--T+V&-2.00&-1.69&3/3&0/3\\
&RandV--T+A&-0.88&-0.54&0/3&0/3\\
&RandA/V--Text&-3.80&-3.53&3/3&0/3\\
\addlinespace[3pt]
\multirow{3}{*}{Baichuan}&RandA--T+V&-2.19&-1.19&2/3&0/3\\
&RandV--T+A&-7.55&-5.60&3/3&0/3\\
&RandA/V--Text&-3.68&-3.03&3/3&0/3\\
\addlinespace[3pt]
\multirow{3}{*}{MiniCPM}&RandA--T+V&+1.12&+2.28&0/3&3/3\\
&RandV--T+A&-0.38&+1.23&0/3&0/3\\
&RandA/V--Text&-1.39&-0.89&0/3&0/3\\
\addlinespace[3pt]
\midrule
\multicolumn{6}{c}{\textbf{IEMOCAP}}\\
\midrule
\multirow{3}{*}{Qwen}&RandA--T+V&-1.99&-1.42&3/3&0/3\\
&RandV--T+A&-1.92&-1.79&3/3&0/3\\
&RandA/V--Text&-4.46&-4.21&3/3&0/3\\
\addlinespace[3pt]
\multirow{3}{*}{Baichuan}&RandA--T+V&-1.00&-0.42&0/3&0/3\\
&RandV--T+A&-3.14&-2.66&3/3&0/3\\
&RandA/V--Text&-3.27&-3.10&3/3&0/3\\
\addlinespace[3pt]
\multirow{3}{*}{MiniCPM}&RandA--T+V&+0.63&+0.91&0/3&1/3\\
&RandV--T+A&-1.19&-0.96&3/3&0/3\\
&RandA/V--Text&+1.19&+1.62&0/3&3/3\\
\addlinespace[3pt]
\bottomrule\end{tabular}
\end{adjustbox}
\par\smallskip\footnotesize Full seed-specific effects, confidence intervals, sample sizes and $q$ values for both policies are provided in the accompanying machine-readable tables. Positive effects are not counted as harm.
\end{table}

\clearpage
\subsection{Evaluation References and Runtime Details}\label{sec:reference_runtime}
Within each model, the three-mapping analysis uses the primary Text, T+A, T+V and Full reference predictions. The primary MELD evaluations and additional-mapping runs use the recorded model-specific inference configurations and media-processing procedures.

\begin{table}[H]
\caption{Recorded runtime versions for the primary MELD evaluations and additional-mapping runs.}\label{tab:new_runtime}
\centering\tinytab
\begin{adjustbox}{max width=\linewidth}
\begin{tabular}{lll}\toprule
Model&PyTorch&Transformers\\\midrule
Qwen&2.11.0+cu128&4.57.6\\
Baichuan&2.4.0+cu124&4.45.0\\
MiniCPM&2.3.1&4.44.2\\
\bottomrule\end{tabular}
\end{adjustbox}

\end{table}

Model identifiers, mapping hashes, script hashes and runtime records accompany the machine-readable results. The additional mappings evaluate behavioral robustness; mechanistic results are from the Qwen/MELD cohort described in Section~\ref{sec:mechanistic}.

\FloatBarrier
\section{Failed-Output Sensitivity}\label{app:failures}
We distinguish common-valid estimates from all-target estimates retaining invalid and OOM outcomes as incorrect. The first estimates performance conditional on both outputs being valid; the second also reflects failures to produce a valid label. Effects are utterance-weighted and inference is clustered by dialogue. The accompanying CSV tables report both policies' effects, confidence intervals, $p$, $q$, sample sizes and exclusions.

For the primary seed-42 predictions, the 10,000-resample sensitivity analysis uses families of 18 utility, 18 omission and six replacement comparisons. The primary tables in Sections~\ref{app:replacement} and~\ref{app:paired} use 1,000 resamples with the same underlying predictions. For the three-mapping analysis, families have 18 replacement and 54 omission tests under each policy. BH-FDR adjustment is performed separately within each testing family and failed-output policy. The primary MELD and additional-mapping runs permit one OOM retry with unchanged inputs; final outcomes and retry records are retained rather than deleting failed targets.

\begin{table}[H]
\caption{Primary seed-42 MELD failed-output counts. Every condition has 2,610 expected targets. Invalid and OOM are disjoint final outcomes; other failed-outcome counts are zero.}\label{tab:failures_original_meld}
\centering\tinytab
\begin{adjustbox}{max width=\linewidth}
\begin{tabular}{llrrr}\toprule
Model&Condition&Valid&Invalid&OOM\\\midrule
\multirow{7}{*}{Qwen}&Text&2607&3&0\\
&T+A&2607&3&0\\
&T+V&2606&2&2\\
&Full&2606&2&2\\
&RandA&2606&2&2\\
&RandV&2606&2&2\\
&RandA/V&2606&2&2\\
\addlinespace[3pt]
\multirow{7}{*}{Baichuan}&Text&2610&0&0\\
&T+A&2610&0&0\\
&T+V&2605&0&5\\
&Full&2607&0&3\\
&RandA&2605&0&5\\
&RandV&2608&0&2\\
&RandA/V&2608&0&2\\
\addlinespace[3pt]
\multirow{7}{*}{MiniCPM}&Text&2604&6&0\\
&T+A&2608&2&0\\
&T+V&2595&13&2\\
&Full&2607&1&2\\
&RandA&2604&2&4\\
&RandV&2606&0&4\\
&RandA/V&2601&7&2\\
\addlinespace[3pt]
\bottomrule\end{tabular}
\end{adjustbox}

\end{table}

\begin{table}[H]
\caption{Primary seed-42 IEMOCAP failed-output counts. Every condition has 7,380 expected targets. Invalid and OOM are disjoint final outcomes; other failed-outcome counts are zero.}\label{tab:failures_original_iemocap}
\centering\tinytab
\begin{adjustbox}{max width=\linewidth}
\begin{tabular}{llrrr}\toprule
Model&Condition&Valid&Invalid&OOM\\\midrule
\multirow{7}{*}{Qwen}&Text&7347&33&0\\
&T+A&7343&37&0\\
&T+V&7349&31&0\\
&Full&7355&25&0\\
&RandA&7356&24&0\\
&RandV&7346&34&0\\
&RandA/V&7354&26&0\\
\addlinespace[3pt]
\multirow{7}{*}{Baichuan}&Text&7380&0&0\\
&T+A&7373&7&0\\
&T+V&7373&0&7\\
&Full&7372&0&8\\
&RandA&7373&0&7\\
&RandV&7358&2&20\\
&RandA/V&7373&0&7\\
\addlinespace[3pt]
\multirow{7}{*}{MiniCPM}&Text&7371&9&0\\
&T+A&7376&4&0\\
&T+V&7336&44&0\\
&Full&7369&11&0\\
&RandA&7320&60&0\\
&RandV&7357&23&0\\
&RandA/V&7270&110&0\\
\addlinespace[3pt]
\bottomrule\end{tabular}
\end{adjustbox}

\end{table}

\begin{table}[H]
\caption{Additional-mapping failed-output counts. Invalid and OOM are disjoint final outcomes; other failed-outcome counts are zero. OOM counts denote failures remaining after the prespecified single identical-input retry.}\label{tab:failures_new}
\centering\tinytab
\begin{adjustbox}{max width=\linewidth}
\begin{tabular}{llrrrrrr}\toprule
\multirow{2}{*}{Model}&\multirow{2}{*}{Condition}&\multicolumn{3}{c}{Seed 43}&\multicolumn{3}{c}{Seed 44}\\\cmidrule(lr){3-5}\cmidrule(lr){6-8}&&Valid&Invalid&OOM&Valid&Invalid&OOM\\\midrule
\multicolumn{8}{c}{\textbf{MELD} ($N=2,610$ per condition and seed)}\\
\midrule
\multirow{3}{*}{Qwen}&RandA&2606&2&2&2606&2&2\\
&RandV&2606&2&2&2606&2&2\\
&RandA/V&2607&1&2&2607&1&2\\
\addlinespace[3pt]
\multirow{3}{*}{Baichuan}&RandA&2604&0&6&2604&0&6\\
&RandV&2605&0&5&2606&1&3\\
&RandA/V&2608&0&2&2607&0&3\\
\addlinespace[3pt]
\multirow{3}{*}{MiniCPM}&RandA&2600&6&4&2601&5&4\\
&RandV&2605&1&4&2606&0&4\\
&RandA/V&2599&9&2&2598&10&2\\
\addlinespace[3pt]
\midrule
\multicolumn{8}{c}{\textbf{IEMOCAP} ($N=7,380$ per condition and seed)}\\
\midrule
\multirow{3}{*}{Qwen}&RandA&7360&20&0&7360&20&0\\
&RandV&7344&36&0&7345&35&0\\
&RandA/V&7352&28&0&7357&23&0\\
\addlinespace[3pt]
\multirow{3}{*}{Baichuan}&RandA&7374&0&6&7374&0&6\\
&RandV&7372&4&4&7375&1&4\\
&RandA/V&7376&0&4&7376&0&4\\
\addlinespace[3pt]
\multirow{3}{*}{MiniCPM}&RandA&7318&62&0&7330&50&0\\
&RandV&7375&5&0&7368&12&0\\
&RandA/V&7280&100&0&7299&81&0\\
\addlinespace[3pt]
\bottomrule\end{tabular}
\end{adjustbox}
\par\smallskip\footnotesize Across the 179,820 additional predictions, 179,216 are valid, 519 are invalid and 85 remain OOM. Recovered retries are retained in the per-target retry records.
\end{table}

\subsection{Findings Sensitive to the Failed-Output Policy}
Table~\ref{tab:policy_cases} shows the main boundary cases. All 18 replacement effects are negative under both policies. All 18 are significant both on common-valid pairs and on all targets. Statistical significance is assessed using the specified FDR-adjusted sign-flip tests.

\begin{table}[H]
\caption{MiniCPM policy-sensitive omission comparisons from the three-mapping analysis. Effects are in pp; $q$ uses the separate 54-test omission family within each policy. Excluded is the number omitted from common-valid estimation.}\label{tab:policy_cases}
\centering\tinytab
\begin{adjustbox}{max width=\linewidth}
\begin{tabular}{lclrrrrr}\toprule
Dataset&Seed&Comparison&\multicolumn{2}{c}{Common-valid}&\multicolumn{2}{c}{All-target}&Excluded\\\cmidrule(lr){4-5}\cmidrule(lr){6-7}&&&Effect&$q$&Effect&$q$&\\\midrule
IEMOCAP&42&RandA/V--Text&+1.56&.00389&+0.98&.07573&119\\
IEMOCAP&43&RandA/V--Text&+1.62&.00282&+1.06&.04976&108\\
IEMOCAP&44&RandA/V--Text&+1.19&.02640&+0.79&.13302&89\\
\bottomrule\end{tabular}
\end{adjustbox}

\end{table}

In the 54-test common-valid omission family, 43 effects are negative and 11 positive; 32 negative and seven positive effects have $q<.05$. The all-target family has 43 negative, ten positive and one zero effect; 32 negative and five positive effects have $q<.05$. MiniCPM/IEMOCAP RandA/V-minus-Text is positive and significant for all three common-valid comparisons, but only seed 43 remains significant on all targets ($q=.04976$).

For MiniCPM on MELD, the primary seed-42 comparison of RandV relative to T+A has common-valid $q=.04962$ and all-target $q=.07139$ within its 18-test omission family. This is distinct from the three-mapping 54-test result; the corresponding 1,000-resample estimate in Section~\ref{app:paired} has $q=.0599$.

\subsection{Metric Denominators}
Accuracy and Weighted F1 in Table~\ref{tab:main}, and Macro-F1 and Balanced Accuracy in Table~\ref{tab:robust}, are computed over all expected targets for both datasets. Invalid and OOM outcomes contribute no true positives and remain in gold-class support. The supplementary metric-policy table reports values under both all-target and valid-only denominators.

\FloatBarrier
\clearpage

\end{document}